\documentclass[10pt]{article}

\usepackage[accepted]{icml_codeml_2025}

\usepackage{xcolor}

\usepackage{graphicx}
\usepackage{subcaption}
\usepackage{booktabs} 
\usepackage{microtype}

\usepackage{hyperref}

\usepackage{multirow}
\usepackage{makecell}
\usepackage{wrapfig} 
\usepackage{enumitem}

\usepackage{pifont}

\usepackage{amsmath}
\usepackage{amssymb}
\usepackage{mathtools}
\usepackage{amsthm}
\usepackage{ulem}
\usepackage{algpseudocode}
\usepackage{listings}

\usepackage[capitalize]{cleveref}

\theoremstyle{plain}
\newtheorem{theorem}{Theorem}[section]

\theoremstyle{definition}
\theoremstyle{remark}

\usepackage[disable,textsize=tiny]{todonotes}

\newif\ifshowcomments

\ifshowcomments
\newcommand{\thib}[1]{{\color{blue}[thib]: #1}}
\newcommand{\agus}[1]{{\color{blue}[agus]: #1}}
\newcommand{\fel}[1]{{\color{pink}[fel]: #1}}
\newcommand{\franck}[1]{{\color{red}[franck]: #1}}
\newcommand{\va}[1]{{\color{yellow}[val]: #1}}
\else
\newcommand{\thib}[1]{}
\newcommand{\agus}[1]{}
\newcommand{\fel}[1]{}
\newcommand{\franck}[1]{}
\newcommand{\va}[1]{}
\fi

\newcommand{\Orthogonium}{ \href{https://github.com/deel-ai/orthogonium/}{Orthogonium }}

\newcommand{\mathbconv}{\circledast} 
\newcommand{\kernel}[1]{\mathbf{#1}}

\newcommand{\toep}[1]{\mathcal{#1}}

\newcommand{\convker}[1]{\star_{#1}}

\icmltitlerunning{\Orthogonium: A Unified, Efficient Library of Orthogonal and 1‑Lipschitz Building Blocks}

\begin{document}

\twocolumn[

\icmltitle{\Orthogonium: A Unified, Efficient\\
       Library of Orthogonal and 1‑Lipschitz Building Blocks}
\begin{icmlauthorlist}
\icmlauthor{Thibaut Boissin}{irt,aniti,irit}
\icmlauthor{Franck Mamalet}{irt,aniti}
\icmlauthor{Valentin Lafargue}{irt,aniti,nowimt}
\icmlauthor{Mathieu Serrurier}{irit,aniti}
\end{icmlauthorlist}

\icmlaffiliation{irt}{Institut de Recherche Technologique Saint-Exupéry, Toulouse, France}
\icmlaffiliation{aniti}{Artificial and Natural Intelligence Toulouse Institute, France}
\icmlaffiliation{irit}{IRIT, Toulouse, France}
\icmlaffiliation{nowimt}{now at IMT, Toulouse, and INRIA, Bordeaux, France}

\icmlcorrespondingauthor{Thibaut Boissin}{thibaut.boissin@irt-saintexupery.com}

\icmlkeywords{Orthogonal convolution, 1-Lipschitz, Open Source, Robust Machine Learning, ICML}
\vskip 0.3in
]

\printAffiliationsAndNotice{}
\begin{abstract}
Orthogonal and 1-Lipschitz neural network layers are essential building blocks in robust deep learning architectures, crucial for certified adversarial robustness, stable generative models, and reliable recurrent networks. Despite significant advancements, existing implementations remain fragmented, limited, and computationally demanding. To address these issues, we introduce \textbf{\Orthogonium}, a unified, efficient, and comprehensive PyTorch library providing orthogonal and 1-Lipschitz layers. \Orthogonium provides access to standard convolution features—including support for strides, dilation, grouping, and transposed-while maintaining strict mathematical guarantees. Its optimized implementations reduce overhead on large scale benchmarks such as ImageNet. Moreover, rigorous testing within the library has uncovered critical errors in existing implementations, emphasizing the importance of standardized and reliable tools. \Orthogonium thus significantly lowers adoption barriers, enabling scalable experimentation and integration across diverse applications requiring orthogonality and robust Lipschitz constraints.
Orthogonium is available \href{https://github.com/deel-ai/orthogonium/}{here}.

\end{abstract}

\section{Introduction}
\label{sec:intro}

    1-Lipschitz neural networks constrain transformations to preserve input norms, providing tight, certifiable robustness against adversarial attacks~\cite{szegedy_intriguing_2013, anil_sorting_2019}. Orthogonal layers reinforce the 1-Lipschitz constraint by requiring an exact unity constant in almost all directions, providing tighter global certification guarantees. Besides robustness, these layers benefit normalizing flows~\cite{kingma2018glow,behrmann2019invertible}, Wasserstein GANs~\cite{arjovsky2017wasserstein,gulrajani2017improved}, stable recurrent architectures~\cite{kiani_projunn_2022,qi_deep_isometric_2020,bansal2018gainorthogonalityregularizationstraining}, and physics-informed models. \va{Cite for Lipschitz PINN? Demander à Paul?}

    \begin{figure}[t]
        \centering
        \includegraphics[width=0.99\linewidth]{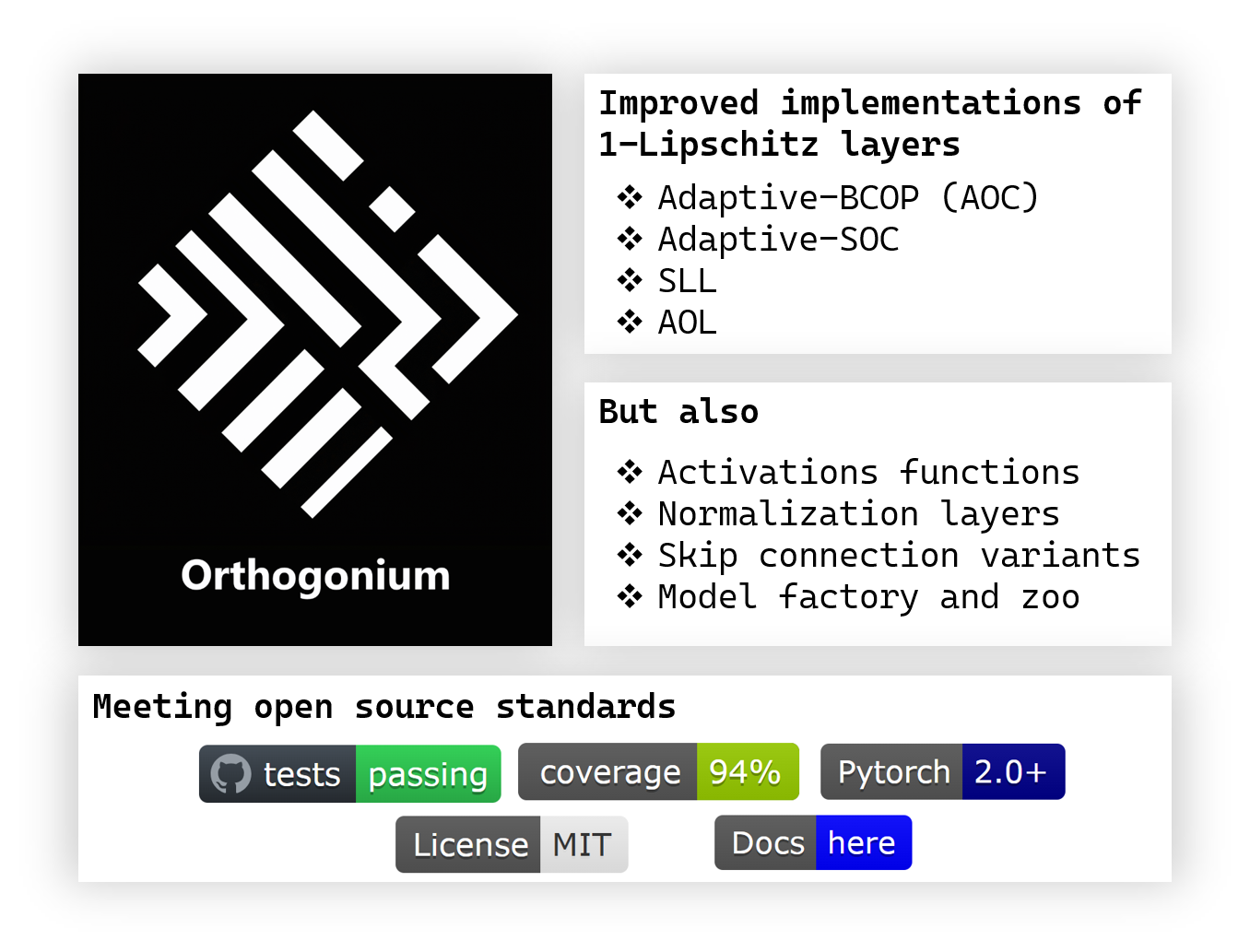}
        \caption{\Orthogonium offers a standardized API to use, and create 1-Lipschitz layers, allowing a user to construct, test, and improve easily such a kind of network}
        \label{fig:bigpic}
    \end{figure}
    
    \paragraph{Motivations.}
    Over the last decade, the certifiable‑robustness community has produced an impressive toolbox of 1‑Lipschitz building blocks—orthogonal and norm‑controlled layers, specialized activations, residual schemes, and normalization layers. Unfortunately, these ingredients remain scattered across dozens of papers and repositories. For practitioners who simply “need a 1‑Lipschitz backbone’’—be they working on Wasserstein GANs, stable RNNs, privacy‑preserving analytics, or safety‑critical perception—the landscape is opaque: Which methods are still maintained? Which supports modern CNN staples such as stride, dilation, grouping, or transposed convolutions? Which combinations scale to ImageNet?
    
    Constructing a truly 1‑Lipschitz network exacerbates the problem. Every layer in every branch must respect the global constraint, yet many recent proposals cover only the vanilla $3\times3$ convolution and ignore grouped, dilated, or strided variants that dominate contemporary architectures \cite{liu2022convnet, tan2019efficientnet, sandler2018mobilenetv2, ronneberger2015u}. In practice, researchers resort to copy‑pasting the original authors’ code—sometimes years out of date—because re‑implementing and validating the underlying mathematics (let alone optimizing kernels) is prohibitively time‑consuming. Convolutional layers are a prime example: half a dozen orthogonalization schemes exist, but none has achieved field‑wide consensus, and their relative merits are hard to benchmark because no common interface or test‑bed exists.
    
    Training itself is also expensive. Certifiable objectives typically require longer schedules to converge, and per‑batch cost grows as soon as weights are iteratively projected or parameter matrices inflated for numerical stability. Here, implementation details matter: an efficient CUDA kernel or memory‑lean fusion can translate directly into larger batches, deeper models, or simply more optimization steps on the same hardware budget.
    
    These main points motivate \textbf{\Orthogonium}. By \textit{centralizing} published methods behind a unified PyTorch API, \textit{standardizing} their signatures and test coverage, and prioritizing \textit{efficient, scalable} kernels, the library (i) turns method comparisons into one‑liner swaps, (ii) lowers the entry-barriers for neighboring fields to adopt 1‑Lipschitz layers, and (iii) makes large‑scale experiments—ImageNet‑1K or semantic segmentation tasks—practically feasible.
    
    \paragraph{Our Contributions.}
    To address these challenges, we introduce \textbf{\Orthogonium}, a unified, efficient library that combines theoretical rigor and practical implementation. Our main contributions are:
    \begin{itemize}
    \item \textbf{Unified, Explicit API:} A comprehensive, PyTorch-friendly implementation covering dense, convolutional, and hybrid orthogonal layers, explicitly constructed in the spatial domain for straightforward integration.
    \item \textbf{Full Feature Parity:} Native support for essential convolutional operations—striding, dilation, transposition, and grouping—allowing seamless integration into modern network architectures.
    \item \textbf{Efficient and Scalable Implementation:} Optimized kernels provide high performance, with minimal overhead (approximately 10\% slowdown) compared to unconstrained convolutions on large-scale benchmarks like ImageNet \cite{deng2009imagenet}.
    \item \textbf{Cross-Fertilization and Flexibility:} \Orthogonium provides modularity to swiftly explore hybrid approaches, enhancing existing methods such as SOC, SLL, and Sandwich layers, promoting broader adoption.
    \item \textbf{Extensive Validation and Testing:} Our rigorous unit testing identified and corrected subtle implementation errors in published repositories, improving reliability and correctness across all supported methods.
    \end{itemize}
    
    By unifying orthogonality, flexibility, and computational efficiency, \Orthogonium represents a significant advancement, enabling researchers and practitioners to integrate orthogonal layers seamlessly into a wide variety of deep learning applications. The remainder of the paper is structured as follows: \cref{sec:dense} introduces our dense layer implementations, \cref{sec:conv} covers orthogonal convolutions, and \cref{sec:unit_testing} covers the approach we used to unit test all our layers. The issues identified in certain approaches underscore the importance of open source tools for safety-critical applications.
    
\section{Dense Orthogonal Layers}
\label{sec:dense}

    \Orthogonium provides an efficient and flexible implementation of orthogonal dense layers with a straightforward, drop-in PyTorch interface, \verb|OrthoLinear|, supporting several orthogonalization methods. This approach simplifies integration into existing workflows and allows users to choose methods suited to their computational constraints and stability requirements. Below, we outline the supported methods and their characteristics:
    
    \paragraph{Unified API with \texttt{OrthoLinear}.}
    The provided \verb|OrthoLinear| class extends \verb|torch.nn.Linear|, ensuring seamless integration into standard PyTorch models. It enforces orthogonality constraints through parameterizations registered via the customizable \verb|OrthoParams| object, which encapsulates both spectral normalization and orthogonalization methods.
    
    \paragraph{Supported Orthogonalization Methods.}
    \Orthogonium supports five distinct orthogonalization algorithms, each appropriate for different scenarios: Björck–Bowie Iterative Projection, Exponential Map method, Modified Gram–Schmidt (QR Decomposition), Cayley Transform, Cholesky Decomposition. These parametrizations are fully compatible with the PyTorch \texttt{parametrize} API.

        
        
        
    
    \paragraph{Spectral Normalization.}
    To ensure numerical stability and enforce Lipschitz constraints, spectral normalization via batched power iteration is applied optionally before orthogonalization. This preconditioning enhances the stability and convergence of the orthogonalization processes. Spectral normalization can also be used in a standalone way, leading to 1-Lipschitz layers.
    
    \paragraph{Flexibility and Extensibility.}
    Users can easily customize orthogonalization and normalization methods through the \verb|OrthoParams| object. \Orthogonium provides several predefined configurations depending on the desired method.
    
    \paragraph{Implementation Efficiency.}
    Implementation and efficiency are crucial factors in the selection of a layer. This is why \Orthogonium provides layers with some non-trivial modifications in order to be more scalable. 
    
    By providing a unified API and efficient implementations, \Orthogonium’s dense orthogonal layers enable easy integration, rigorous validation, and high-performance execution in diverse deep learning applications.

\section{Orthogonal Convolutions}
\label{sec:conv}
    
    \Orthogonium implements multiple classes of 1-Lipschitz and orthogonality-preserving
    convolutions—allowing for a user to choose between exactness,
    speed, and architectural flexibility— plus two 1-Lipschitz extensions
    that embed the convolutions inside higher-order residual blocks.
    Table~\ref{tab:conv_summary} summarizes their properties, while the paragraphs below describe design choices and implementation tweaks; algorithmic derivations are deferred to \cref{sec:technical_details}. As the exposed layers can differ significantly from their original implementations, original layers are available in the \texttt{legacy} module.
    \paragraph{Adaptive Orthogonal Convolution (AOC).}
    
    \textbf{AOC} is the default constructor, \verb|AdaptiveOrthoConv2d|/\verb|ConvTranspose2d|, and generalizes BCOP kernels~\cite{li_preventing_2019} to \emph{any} kernel size, stride, dilation, groups natively (i.e, without resorting to     reshaping tricks or FFTs). Transposed convolutions are also supported natively. It is based on the method described in \cite{boissin2025adaptive}.  The layer materializes an explicit weight tensor whose forward path is a single call to \verb|torch.nn.Conv2d| This approach yields a \(\le1.13\times\) wall-time over plain \verb|Conv2d| on ImageNet-1k at batch size~256.
       
    \paragraph{Adaptive-SOC.}
    Skew Orthogonal Convolution (SOC) offers orthogonal training by
    parameterizing the kernel as the exponential of a skew-symmetric
    filter~\cite{singla_skew_2021}.  \Orthogonium’s
    \verb|AdaptiveSOCConv2d|/\verb|ConvTranspose2d| fuses SOC with AOC’s
    stride-aware approach, stores the \emph{explicit} exponential once per
    update (making its cost independent of the batch size), and supports grouped, dilated or
    transposed variants out-of-the-box—reducing memory (See \cref{sec:improving_soc}). Also, this method relies on a normalization step. We used "AOL" instead of the original "fantastic four"\cite{singla_fantastic_2021} approach, making the convergence quicker than the original method. (3-4 iterations instead of the original 6).
     
    \paragraph{Almost-Orthogonal Layers (AOL).}
    When strict orthogonality is unnecessary, \verb|AOLConv2d| implements the almost-orthogonal method of Prach \& Lampert \cite{prach_almost-orthogonal_2022}.  The
    re-parametriser is registered through PyTorch’s
    \verb|parametrize| API and guarantees a layer Lipschitz constant
    \(\le1\), while remaining fast. \Orthogonium 
    implements a multi-step variant, making use of the proximity between this approach and the Gram 
    iteration described by \cite{delattre2024spectral}. This variant allows a tighter normalization than the original method.
    
    \franck{voir ma remarque de separer ces layers ds une autre section,surtout que tu enchaine sur les activation, and co...}\thib{Je suis d'accord, mais j'ai pas réussi à trouver un bon moyen d'intégrer SLL au paragraph "residual" de manière fluide en temps et en heure :-)}
    
    \paragraph{SDP-based Lipschitz Layers (SLL).}
    \cite{araujo_unified_2023} defined a 1-Lipschitz residual blocks that bundle a
    \(\sigma(\cdot)\)-non-linearity inside the convolution, while offering a tight Lipschitz normalization. We extended the original
    \verb|SDPBasedLipschitzConv| to support groups and dilation. We also designed its down-sampling equivalent
    \verb|SLLxAOCLipschitzResBlock| which allows for stride and dimension change with an AOC kernel at their core to enable strides and channel changes, similarly as the Resnet downsampling blocks (\cref{fig:SLL}); details are given in
    \cref{sec:improving_sll}.
    
    \paragraph{Sandwich-AOC.}
    Finally, \Orthogonium replaces the costly frequency-domain Cayley step of
    “sandwich layers’’ \cite{wang_direct_2023} with an explicit
    AOC kernel, removing complex-valued FFTs, making this layer efficient for large input images (e.g, $224 \times 224 $ ); details are given in
    \cref{sec:improving_sandwich}. This layer is still under development and will be available soon.

    \begin{figure*}
        \centering
    \caption{Implemented convolutional layers in \Orthogonium.  All run on
    GPU, accept \texttt{stride}, \texttt{dilation}, \texttt{groups},
    \texttt{padding\_mode}, and have (when possible) parity with \texttt{nn.Conv2d}.}
    \label{tab:conv_summary}
    \small
    \begin{tabular}{lccc}
    \toprule
    Layer & Orthogonality & Key use-case & Internal method\\
    \midrule
    AOC (\texttt{AdaptiveOrthoConv2d}) & exact & general CNN backbones & lifted BCOP / RKO\\
    Adaptive-SOC & exact & depthwise / small kernel size & exponential skew filter\\
    AOL & $\le1$-Lip.\,($\approx$) & fast training & multi-step projection\\
    SLL / SLL-AOC & $\le1$-Lip.\,(tight) & residual blocks & AOL + SDP constraint\\
    Sandwich-AOC & $\le1$-Lip.\,(tight) & tight Lipschitz estimation without orthogonality & AOC pair\\
    \bottomrule
    \end{tabular}
    \end{figure*}

\paragraph{Other Modules: Activations, Normalization, and Residual Blocks.}
Beyond convolutions, \textbf{\Orthogonium} centralises the necessary layers needed to build \emph{fully} $1$-Lipschitz networks.

\textbf{Activations.}  
Five gradient-norm-preserving non-linearities are shipped in \texttt{custom\_activations}: \texttt{Abs}, \texttt{SoftHuber}, \texttt{MaxMin}, \texttt{HouseHolder}, and its second-order variant.  Each is unit-Lipschitz by construction; for the Householder family we patched the missing $1/\!\sqrt{2}$ scale factor, restoring $\sigma_{\max}(J)=1$ to machine precision (see \cref{ap:unit_testing}).  Detailed APIs appear in the online docs.

\textbf{Normalization.}  
Instead of BatchNorm—which destroys Lipschitz control—\Orthogonium offers \texttt{BatchCentering} and \texttt{LayerCentering}.  
Both subtract running means but leave variances untouched, so they preserve feature-map norms at inference; their implementation lives in \texttt{normalization.py}.


Since the usual residual connection is not 1-Lipschitz, \Orthogonium provides several lightweight residual wrappers designed to combine an arbitrary internal function $\mathrm{fn}$—typically an orthogonal convolution followed by non-linear activation—with a skip connection, ensuring that the resulting residual blocks remain exactly $1$-Lipschitz. These wrappers, implemented as minimal `torch.nn.Module` classes, preserve the benefits of skip connections while strictly enforcing global Lipschitz constraints.




Among the various strategies, \textit{ConcatResidual} splits the input along the channel dimension, applies a function $\mathrm{fn}$ to one half, and concatenates it back with the untouched half, ensuring the resulting block remains $1$-Lipschitz if $\mathrm{fn}$ is. \textit{L2NormResidual} combines the identity and residual branches using an $\ell_{2}$ average, specifically outputting $\sqrt{\tfrac12 x^{2}+\tfrac12\mathrm{fn}(x)^{2}+\varepsilon}$ (with small $\varepsilon$ for numerical stability), ensuring exact $1$-Lipschitz continuity. \textit{AdditiveResidual} and its variation, \textit{PrescaledAdditiveResidual}, form convex combinations of the identity and transformed branches using a learnable scalar gate $\alpha$: the former via interpolation $\alpha x + (1-\alpha)\mathrm{fn}(x)$ (with $\alpha$ constrained by a sigmoid), and the latter by premultiplying the input as $ \frac{x+\mathrm{fn}(\alpha x)}{1+\lvert\alpha\rvert}$, where $\alpha$ unconstrained.




\section{Unit testing of constrained layers} \label{sec:unit_testing}

     Despite theoretical guarantees, empirical verification remains crucial to detect (i) numerical instabilities (e.g., floating-point precision errors) and (ii) implementation discrepancies (e.g., padding mismatch, missing factor), ensuring orthogonality in practice. \va{ Parler rapidement de ce qui a été fait en dense ? / Dense pas de problème donc moins de test nécessaire}\thib{Cette phrase englobe dense/conv/non linear il y a quelques centaines de tests en dense aussi :-p} More details about our unit testing scheme are available in \cref{ap:unit_testing}.

    \paragraph{Unit testing of convolutional layers.}
    Orthogonality depends on training hyperparameters and convolution parameters (stride, group, dilation, transposition). We ensure correctness by combining two methods: (i) explicit SVD on Toeplitz matrices for precise validation on small inputs, and (ii) scalable spectral methods (from \texttt{conv.singular\_values}) for practical, large-scale validation (as it uses parametrization-aware optimizations). All tests maintain singular value tolerance ranging from $10^{-4}$ to $5e^{-3}$ (for some methods).
    
    We used both of these approaches in our unit tests. This enables us to ensure that the second method (which is faster and more scalable) is correct and effectively checks for layer orthogonality. We also added several unit tests to ensure that impossible theoretical configurations—as described in \cite{achour2022existence}—are rejected.

    \paragraph{Unit testing of non-linear layers.}
    To guarantee that activations and higher-order residual blocks do not silently violate the global $1$-Lipschitz constraint, we complement the linear checks described above with a non-linear test suite based on the empirical Jacobian (computed with automatic differentiation) computed on randomly sampled and optimized tensors. This verified that the optimizer updates kept the block weights within the desired constraints.

    \paragraph{Issues uncovered using unit-testing}
    Crucially, these tests uncovered a flaw in the original \emph{HouseHolder} activation \cite{singla_improved_2022}: the reflection missed a $1/\sqrt{2}$ normalization and was therefore $\sqrt{2}$-Lipschitz. Re-scaling the kernel collapses all singular values to $1\pm10^{-5}$, after which the layer satisfies our criteria.  
    
    Overall, this test bank was of precious use to confirm that all parameters can be combined in practice (i.e., A strided-grouped-transposed convolution with a dilation factor is still orthogonal). The library achieves 94\% test coverage overall.


\section{Conclusion}

\Orthogonium unifies a decade of advances in orthogonal and Lipschitz-constrained layers into a single, efficient, and comprehensive PyTorch library. By providing native support for strided, dilated, grouped, and transposed convolutions—alongside rigorous validation and optimized code—it significantly reduces implementation overhead, fosters reliable experimentation, and promotes adoption in critical applications such as certified robustness, generative modeling, and stable recurrent architectures. Furthermore, open-sourcing \Orthogonium facilitates rigorous testing and validation by the broader community, uncovering subtle implementation errors and enabling ongoing verification and improvements. \Orthogonium thus serves as a foundational resource, bridging theory and practice to enable scalable, robust, and provably stable deep learning architectures.

\section*{Acknowledgements}
This work was carried out within the DEEL project,\footnote{\url{https://www.deel.ai/}} which is part of IRT Saint Exupéry and the ANITI AI cluster. The authors acknowledge the financial support from DEEL's Industrial and Academic Members and the France 2030 program – Grant agreements n°ANR-10-AIRT-01 and n°ANR-23-IACL-0002.

\bibliography{full.bib}
\bibliographystyle{icml2025}

\appendix

\newpage

\section{Orthogonalization methods available for 2D weights matrices}
\label{ap:matrix_orth_methods}
\subsection{QR Factorization via Modified Gram–Schmidt}

We implement the Modified Gram–Schmidt (MGS) algorithm \cite{LaPlace1820} for numerically stable QR factorization.  Starting from \(W=[w_i]_{i=0}^{C-1}\in\mathbb{R}^{C\times C}\), we orthogonalize one column at a time, correcting rounding errors at each step:


To enforce a unique factorization and improve stability, we post-multiply \(Q\) by \(\operatorname{sign}(\operatorname{diag}(R))\) so that all diagonal entries of \(R\) are positive.

\subsection{Cayley Transform}

The Cayley transform \cite{Cayley_1846} maps any skew-symmetric \(A\) (\(A^T=-A\)) to an orthogonal \(Q\) via:
\begin{align*}
    Q = (I - A)(I + A)^{-1}
\end{align*}
For rectangular \(W\in\mathbb{R}^{M\times C}\,(M\ge C)\), we follow the partitioning and “augmented” Cayley of \cite{pauli2023lipschitz}:

\begin{algorithm}[H]
\begin{algorithmic}[1]
\Require \(W\in\mathbb{R}^{M\times C}\)
\Ensure \(\hat W\in\mathbb{R}^{M\times C}\), orthogonal columns
\Procedure{Cayley Transform}{ W }
    \State Partition \(W = \begin{bmatrix}U\\V\end{bmatrix}\), \(U\in\mathbb{R}^{C\times C}\)
    \State \(A \gets U - U^T + V^T V\) \Comment{not strictly skew but yields correct block}
    \State \(B \gets (I + A)^{-1}\)
    \State \(\hat W_1 \gets B\,(I - A)\), \(\hat W_2 \gets -2\,V\,B\)
    \State \(\hat W \gets [\,\hat W_1;\,\hat W_2\,]\)
    \State \Return \(\hat W\)
\EndProcedure
\end{algorithmic}
\caption{Augmented Cayley Transform}
\label{alg:cayley}
\end{algorithm}

Matrix inversion can be a bottleneck, so in practice we cache and reuse factorizations where possible.

\subsection{Exponential Map}

Using the Lie-group exponential \(\exp(A)\) of a skew-symmetric \(A=W-W^T\) yields an orthogonal matrix since \(\exp(A)^T = \exp(-A)\).  We normalise \(A\) by its spectral norm to avoid overflow and truncate the power series after \(p\) terms \cite{singla_skew_2021}:

\begin{algorithm}[H]
\begin{algorithmic}[1]
\Require \(W\in\mathbb{R}^{C\times C}\), \(p\in\mathbb{N}\)
\Ensure \(\hat W = \exp(A)\) with \(A=-A^T\)
\Procedure{Lipschitz Exponential}{W,p}
    \State \(A \gets W - W^T\)
    \State \(\hat A \gets A / \|A\|_2\) \Comment{spectral normalization}
    \State \(\hat W \gets I_C\), \(\hat A_k \gets I_C\)
    \For{\(k=1,\dots,p\)}
        \State \(\hat A_k \gets \frac1k\,\hat A_k\,\hat A\)
        \State \(\hat W \gets \hat W + \hat A_k\)
    \EndFor
    \State \Return \(\hat W\)
\EndProcedure
\end{algorithmic}
\caption{Exponential Map with Spectral Normalization}
\label{alg:EXP}
\end{algorithm}

This method parametrizes only \(\mathrm{SO}(C)\) and its accuracy depends on \(p\).

\subsection{Cholesky Decomposition}

Following \cite{hu_recipe_2023}, we form the Gram matrix \(C = W W^T + \varepsilon I\) (\(\varepsilon>0\) for PD), compute its Cholesky \(C = L L^T\), and solve
\[
  L\,\hat W = W \quad\Longrightarrow\quad \hat W = L^{-1} W,
\]
which enforces \(\hat W\,\hat W^T = I\).  This triangular solve is \(O(C^3)\) but highly tuned in modern BLAS libraries.

\subsection{Björck–Bowie Iterative Projection}

The Björck–Bowie algorithm \cite{bjorck1971iterative} finds the nearest orthogonal matrix by fixed-point iteration
\[
  W_{t+1} = (1+\beta)\,W_t - \beta\,W_t W_t^\top W_t,
  \quad \beta\in\bigl(0,\tfrac12\bigr].
\]
With spectral normalization of \(W_0\) and \(\beta=\tfrac12\), convergence is fast: 12–25 iterations suffice in practice \cite{anil_sorting_2019}.  We reuse a cached power-iteration estimate of the top singular vector across updates to further accelerate each step.

\section{Technical details of improved orthogonal convolution methods}\label{sec:technical_details}

    \thib{introduce Block convolution and usages}

    \subsection{Improving BCOP (AOC)\cite{boissin2025adaptive}}

    \Orthogonium introduces \textit{Adaptive Orthogonal Convolution} (AOC),\cite{boissin2025adaptive}, combining the strengths of BCOP \cite{li_preventing_2019} and Reshaped Kernel Orthogonalization (RKO) \cite{serrurier_achieving_2021}. BCOP constructs explicit orthogonal convolution kernels by composing elementary orthogonal building blocks (such as $1\times 1$, $1\times 2$, and $2\times 1$ convolutions), but originally lacks support for advanced convolutional operations like stride, dilation, transposition, or grouped convolutions. Conversely, RKO supports native striding but typically achieves only approximate orthogonality.
    
    AOC addresses these limitations by integrating BCOP and RKO into a single orthogonal convolutional kernel. Specifically, given a desired stride $s=k$, AOC defines the convolutional kernel as:
    \begin{equation}
    {\text{AOC}}= {\text{RKO}} \mathbconv \kernel{K}_{\text{BCOP}}
    \end{equation}
    
    The resulting kernel maintains strict orthogonality and explicitly supports stride, dilation, grouping, and transposed convolutions natively. Crucially, orthogonality is preserved through a careful choice of internal channel dimensions, ensuring both flexibility and computational efficiency \cite{boissin2025adaptive}.
    
    AOC is rigorously proven orthogonal for any valid configuration ($k \ge s$), \cite{achour2022existence}, offering significant practical advantages over existing methods that rely on computationally expensive reshaping or Fourier-based operations.
        
    \paragraph{Native Strided Convolution.} Unlike prior methods that emulate striding via tensor reshaping—leading to substantial computational overhead—AOC implements native striding. This approach avoids the exponential computational complexity of reshaped methods, making orthogonal convolutions feasible for large-scale applications.
    
    \paragraph{Native Transposed Convolution.} By explicitly constructing orthogonal kernels, AOC naturally supports transposed convolutions, crucial for architectures requiring learnable upsampling such as U-Nets \cite{ronneberger2015u} and Variational Autoencoders (VAEs) \cite{kingma2013auto}.
    
    \paragraph{Native Grouped Convolution.} AOC efficiently supports grouped convolutions, widely used in contemporary models such as EfficientNet and ResNeXt, by independently orthogonalizing groups within the convolutional layer.
    
    \paragraph{Dilation.} Orthogonality under dilation follows naturally from AOC’s explicit kernel construction, providing enlarged receptive fields without additional parameter overhead or loss of orthogonality.
    
    These native implementations allow AOC to maintain a minimal overhead (approximately 10\%) compared to unconstrained convolutional models, even at ImageNet scales.

    In this section, we will explore how the content of this paper can be used to improve existing layers from the state of the art.
    
    \subsection{Improving skew orthogonal convolution (SOC)\cite{singla_improved_2022}
    \label{sec:improving_soc}
    } This method, introduced by \cite{singla_improved_2022} uses the fact that an exponential of a skew-symmetric matrix is orthogonal. The initial implementation builds a skew-symmetric kernel and computes the exponential convolution. However, without proper tools to compute the exponential of a convolution kernel, this exponential was computed implicitly for each input by using the Taylor expansion of the exponential (see \cref{eq:soc_implicit}).

        \begin{theorem}[Explicit conv exponential]

            We can use the block convolution operator \footnote{Block convolution operator allows to fuse the kernels of two convolutions to construct the kernel of a convolution equivalent to the composition of the two convolutions. It is defined in \cite{li_preventing_2019}, and an efficient implementation is available in \cite{boissin2025adaptive}} to compute explicitly the exponential of a kernel $\kernel{K}$:

            \begin{align}
                & x + \frac{\kernel{K}*x}{1!} + \frac{\kernel{K}*\kernel{K}*x}{2!} + \dots \label{eq:soc_implicit} \\
                = & \left(Id + \kernel{K} + \frac{\kernel{K} \mathbconv \kernel{K}}{2!} + \frac{\kernel{K} \mathbconv \kernel{K} \mathbconv \kernel{K} }{3!} + \dots \right) * x \label{eq:soc_explicit}            
            \end{align}
        \end{theorem}

         \Cref{eq:soc_explicit} shows that we can compute the exponential of a convolution kernel a single time, while the formulation in \cref{eq:soc_implicit} needs to be done for each input $x$. In other words, we can apply one conv instead of $n_{iter}$ convs. Note that the resulting kernel is then larger than the original one. In theory, this could unlock large speedups, but the gain is limited in practice as the implementation of convolution layers is optimized for small kernels and large images \cite{ding_scaling_2022}. However, the original implementation requires the storage of $n_{iter}$ maps, whereas our implementation only one. This, in practice, unlocks larger networks and batch sizes.

        Also, it is possible to handle a change in the number of channels and striding using a similar approach as AOC layers.

    \subsection{Improving SDP-based Lipschitz Layers (SLL)
    \cite{araujo_unified_2023}
    }
    \label{sec:improving_sll}
    \begin{figure}[h]
        \centering
        \includegraphics[width=0.9\linewidth]{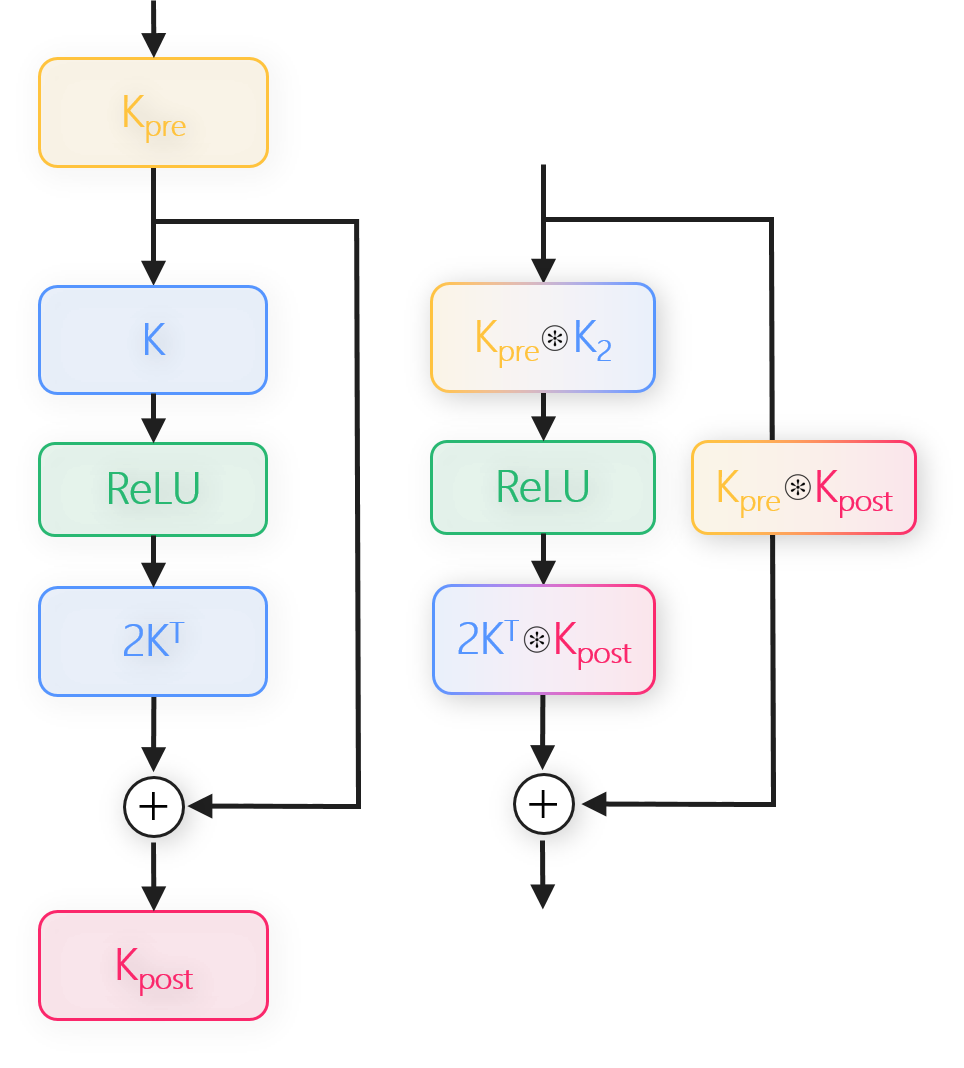}
        \caption{\textbf{The $\mathbconv$ can be used to enable $s\neq 1$ and $c_i \neq c_o$ configurations on SLL.} The flexibility of the $\mathbconv$ allows for operations resulting in a block with a similar structure as the original ResNet block.}
        \label{fig:SLL}
    \end{figure}

    SLL layer for convolutions, proposed in \cite{araujo_unified_2023}, is a 1-Lipschitz layer defined as:
    \[
        y = x - 2\kernel{K}^T \convker{} ( \sigma ( \kernel{K} \convker{} x + b ))
    \]
    Note that in the original paper, the equation is noted with product of two matrices $WT^{-\frac{1}{2}}$, for convolutions it represents toeplitz matrix, i.e. $WT^{-\frac{1}{2}} =  \toep{K}$.
    
    SLL layer does not natively support neither strides nor changes in the channel size.
    We propose to use the $\mathbconv$ to derive a block, based on SLL, that supports stride and $c_i \neq c_o$, and can replace the strided convolutions of the residual branch in architectures like ResNet.  
    
    A natural first step is to append a strided convolution after a SLL block. This layer, $conv_{K_{post}}\circ SLL$, can then be fused in the SLL block thanks to block convolution operator \footnote{as defined in \cite{li_preventing_2019, boissin2025adaptive}}:
    \begin{align*}
        y =& \kernel{K}_{post} \convker{s} (x - 2\kernel{K}^T \convker{} ( \sigma ( \kernel{K} \convker{} x + b ))) \\
        =&  \kernel{K}_{post} \convker{s} x - 2(\kernel{K}_{post} \mathbconv \kernel{K}^T) \convker{s} ( \sigma ( \kernel{K} \convker{} x + b ))) \\
    \end{align*}
    This allows to build a block based on SLL and that supports stride and channel changes. However, this creates an asymmetry between the convolution before the activation and the one after the activation (that has a larger kernel size). 
    
    We propose also to add a second convolution before the SLL block, $conv_{K_{post}}\circ SLL\circ conv_{K_{pre}}$allowing better control over the kernel size of each convolution:
    \begin{align*}
        y = &\kernel{K}_{post} \convker{s} \kernel{K}_{pre} \convker{} x \\
        & - 2(\kernel{K}_{post} \mathbconv \kernel{K}^T) \convker{s} ( \sigma ( \kernel{K} \convker{} \kernel{K}_{pre} \convker{} x + b ))) \\
        = &(\kernel{K}_{post} \mathbconv \kernel{K}_{pre}) \convker{s} x \\
        & - 2(\kernel{K}_{post} \mathbconv \kernel{K}^T) \convker{s} ( \sigma ( (\kernel{K} \mathbconv \kernel{K}_{pre}) \convker{} x + b ))) \\
    \end{align*}
    
    The proposed block is still a 1-Lipschitz layer (as a composition of 1-Lipschitz and orthogonal layers), and support efficiently strides and changes of kernel sizes.
    A visual description is provided in \cref{fig:SLL}. This approach is more efficient than the explicit construction that uses 3 distinct convolutions, as kernels are merged once per batch, and intermediate activations of extra convolutions do not need to be stored backward. Typically, when $\kernel{K}$, $\kernel{K}_{pre}$ and $\kernel{K}_{post}$ are $2 \times 2$ convolutions, this results in a residual block with two $3 \times 3$ convolutions in one branch and a single $4 \times 4$ convolution (with stride 2) in the second. This is very similar to transition blocks found in typical residual networks.

    \subsection{Improving Sandwich Layers \cite{wang_direct_2023}}
    \label{sec:improving_sandwich}
    Introduced by \cite{wang_direct_2023}, this approach aims to construct a 1-Lipschitz network globally rather than constraining each layer independently. In practice, this can be done either by (i) adding constraints between layers or (ii) creating layers that incorporate a non-linearity internally (a.k.a. sandwich layers). However, sandwich layers require an orthogonal matrix at their core. For convolutional layers, this is achieved by performing the orthogonalization of the layer in the Fourier domain, as described in the method from \cite{trockman_orthogonalizing_2021} and shown in their Algorithm 1.
    \begin{algorithm}
        \caption{Sandwich convolutional layer (from \cite{wang_direct_2023})}
        \begin{algorithmic}[1]
            \Require $h_\text{in} \in \mathbb{R}^{p \times s \times s}$, $P \in \mathbb{R}^{(p+q) \times q \times s \times s}$, $d \in \mathbb{R}^q$
            \State $\hat{h}_\text{in} \gets \text{FFT}(h_\text{in})$
            \State $\Psi \gets \text{diag}(e^d), \, \begin{bmatrix} \tilde{A} & \tilde{B} \end{bmatrix}^* \gets \text{Cayley}(\text{FFT}(P))$
            \State $\hat{h}[:, i, j] \gets \sqrt{2} \tilde{B}[:, :, i, j] \hat{h}_\text{in}[:, i, j]$
            \State $\hat{h} \gets \text{FFT}(\sigma(\text{FFT}^{-1}(\hat{h}) + b))$
            \State $\hat{h}_\text{out}[:, i, j] \gets \sqrt{2} \tilde{A}[:, :, i, j] \Psi \hat{h}[:, i, j]$
            \State $h_\text{out} \gets \text{FFT}^{-1}(\hat{h}_\text{out})$
        \end{algorithmic}
    \end{algorithm}

    We can leverage AOC to construct the kernel of an orthogonal convolution, replacing the expensive operation performed in the Fourier domain. Thus, we can construct two kernels, $\kernel{A}$ and $\kernel{B}$, with appropriate constraints between the two and apply the rescaling and non-linearity directly in pixel space:
    \[
    h_\text{out} = \sqrt{2} \kernel{A}^\top \convker{} \Psi \sigma \left( \sqrt{2} \Psi^{-1} \kernel{B} \convker{} h_\text{in} + b \right)
    \]

    In practice this is done by constructing an orthogonal kernel with twice the number of channels that is split into two kernels, namely $\kernel{A}$ and $\kernel{
    B}$. This is expected to be more efficient since the use of the Fourier transform is costly for two reasons: first, it necessitates computation with complex values; and second, the cost of the operation depends on the input size, which can be prohibitive in large-scale settings with $224 \times 224$ images. Consequently, our approach can make such a layer more scalable.

    \subsection{Extending Applicability to other methods.} Beyond the previously discussed approaches that show meaningful opportunities for improvement, our method can enhance a wide range of orthogonal convolutional layers. Specifically, we can incorporate our framework into any alternative orthogonal layers, enabling native support for strides in those layers. Furthermore, our approach can unlock features such as grouped convolutions, transposed convolutions, and dilations, broadening its utility and adaptability.

\section{Technical details of our unit testing scheme} \label{ap:unit_testing}

    \paragraph{Evaluating the Lipschitz constant of a network}
    \thib{PAragraph trop long et un peu hors du sujet du papier. Condenser voire supprimer (en ajoutant les citation dans "the need for an empirical evaluation..."}

    Beyond the creation of a constrained layer, the evaluation of the Lipschitz constant of a layer is by itself an active field: early work used fast Fourier transform to evaluate a lower bound of the Lipschitz constant of a convolutional layer with circular padding \cite{sedghi_singular_2019}. This work was later improved with a method that is quicker \cite{senderovich_towards_2022}, supports other types of padding \cite{grishina_tight_2024}, or allows the extraction of a larger part of the spectrum \cite{boroojeny_spectrum_2024}. The work of \cite{delattre2023efficient} \cite{delattre2024spectral} allows us to compute a certifiable upper bound efficiently under different types of padding.
    It is worth recalling that inferring the global Lipschitz constant of a network given the Lipschitz constant of each layer is an NP-Hard problem\cite{virmaux2018lipschitz}. Then, \cite{pauli_lipschitz_2024,fazlyab_efficient_2019,wang2024scalability} aim to tackle using SDP (Semi-definite programming) tools. Our work can also contribute to this issue as the orthogonal layer allows a tighter product bound (ie. bound using the product of the Lipschitz constant of each layer to evaluate the constant of the whole network).

    \paragraph{The need for an empirical evaluation of the Lipschitz constant of layers.}

    Despite the theoretical guarantees ensuring orthogonality in our construction, empirical checks are necessary to confirm implementation correctness. Such verification prevents two types of issues:
    \begin{enumerate}
        \item \textbf{Checking of numerical Instabilities:} Issues arising from floating-point precision, such as those introduced by small epsilon values added to avoid division by zero.
        \item \textbf{Checking for implementation discrepancies:} Differences between mathematical formalism and its translation to popular frameworks (e.g., SOC proofs assume circular padding, while its implementation uses zero padding, it is hard to determine how this difference affects the Lipschitz constant of such a layer).
    \end{enumerate}

    \paragraph{Checking the orthogonality of a layer under stride, group, transposition, and dilation conditions.}

    Orthogonality is sensitive to convolutional parameters such as stride, groups, dilation, and transposition, as well as training hyperparameters like learning rate, weight decay, and orthogonalization iterations. To robustly validate orthogonality, we combine two complementary approaches: (i) explicit singular-value decomposition (SVD) on convolution-induced Toeplitz matrices, ensuring exactness for small-scale inputs, and (ii) scalable spectral norm estimation via \texttt{conv.singular\_values}, suitable for larger-scale practical validation. We thoroughly test diverse configurations—varying kernel sizes, strides, channel dimensions, and padding—ensuring all singular values remain within a stringent tolerance ranging from $10^{-4}$ to $5e^{-3}$ (for some methods).

    The numerical stability and the convergence of an orthogonal layer is dependent on the training hyper-parameters: mainly the number of iterations used in most methods, but the learning rate and weight decay can also play a significant role. We then need an evaluation method that scales along with the convolution and that can be used at the end of each training. On the other hand, as scalable methods can be imperfect, we also need a method that computes very precise bounds without making any assumptions on the layer parameters (like padding, or stride). In order to overcome this, we tested our layers with two distinct methods:
    
    \textbf{Explicit SVD on Toeplitz Matrices:} Using the impulse response approach, we construct the Toeplitz matrix for any padding and stride, allowing direct computation of singular values. This method, though accurate, is computationally expensive for large input images (in spite of full parallelization of the matrix's construction).
    
    \textbf{Evaluation from the \texttt{conv.singular\_values} module:} We use the more scalable methods (like \cite{delattre2023efficient, delattre2024spectral, grishina_tight_2024}) from this module to check that the produced bounds from this module are valid.
    
    We used both of these two approaches in our unit tests. This enables us to ensure that the second method (which is faster and more scalable) is correct to check that our layer is effectively orthogonal. 

    We tested multiple values for kernel size, stride, dilation, input channels, and output channels. For the kernel size, along with standard configurations of $3 \times 3$ and $5 \times 5$ kernels, we also covered cases for $1 \times 1$ kernels and even-sized kernels. For input/output channels, we covered various relevant inequalities (for instance, when $c_o > c_is^2$ as indicated in \cite{achour2022existence}).
    We ran similar tests for transposed convolution (to the extent of what PyTorch allows: notably, circular padding is not supported for transposed convolutions and could not be tested). Also, as the computation of the singular values using the explicit construction of the Toeplitz matrix is quite expensive, we used it on small $8 \times 8$ images; this is also a good way to check for padding issues, as the kernel size is not negligible with respect to the image size. All the checks over the singular values for both methods were done with a tolerance of $1e^{-4}$. 

    \paragraph{Unit testing of non-linear layers.}
    To guarantee that activations and higher-order residual blocks do not silently violate the global $1$-Lipschitz constraint, we complement the linear checks described above with a non-linear test-suite.  
    For every candidate activation we sample small random tensors, build the full Jacobian with, and compute its spectral norm; checking $\max\sigma(J)\le1+10^{-4}$ and—when applicable— $\min\sigma(J)\ge1-10^{-4}$, certifying both $1$-Lipschitzness and orthogonality.  
    A similar strategy is applied to parametrized layers such as SLL, CPL or AOL \cite{araujo_unified_2023, xiao_dynamical_2018, prach_almost-orthogonal_2022} with the difference that before and after ten optimization steps we re-measure their Jacobian’s spectral norm and assert it never exceeds~$1$, ensuring that optimizer updates cannot drift the block outside the desired constraint.  
    
    Because the Jacobian scales quadratically with the number of activations, inputs are restricted to $8 \times 8$ images or $64$-d vectors—small enough for tractable SVD yet large enough to expose implementation bugs. Empirically, this design provokes failures in under three seconds on a laptop GPU and offers a pragmatic alternative to more expensive SDP-based constants.

\end{document}